\documentclass{article}

\pdfoutput=1

\usepackage[preprint, nonatbib]{nips_2019}

\usepackage[utf8]{inputenc} 
\usepackage[T1]{fontenc}    
\usepackage{hyperref}       
\usepackage{url}            
\usepackage{booktabs}       
\usepackage{amssymb}        
\usepackage{amsfonts}       
\usepackage{nicefrac}       
\usepackage{microtype}      

\usepackage{graphicx}
\usepackage{multicol}
\usepackage{ragged2e}
\usepackage{enumitem}
\usepackage{tablefootnote}
\usepackage{bm}
\usepackage{wrapfig}
\usepackage[ruled]{algorithm2e}
\usepackage[caption=false]{subfig}
\usepackage{booktabs}
\usepackage{array}
\usepackage{hyperref}

\usepackage{mathtools}

\DeclarePairedDelimiter\floor{\lfloor}{\rfloor}
\DeclareMathOperator*{\argmin}{arg\,min} 
\DeclareMathOperator*{\argmax}{arg\,max} 
\newcolumntype{L}[1]{>{\raggedright\arraybackslash}p{#1}}

\graphicspath{ {figures/} }

\title{Interpretable Adversarial Training for Text}

\author{
  Samuel A.~Barham\thanks{University of Maryland, https://www.cs.umd.edu/people/sbarham}\\
  Department of Computer Science\\
  University of Maryland\\
  College Park, MD 20742\\
  \texttt{sbarham@terpmail.umd.edu}\\
  \And
  Soheil Feizi\thanks{University of Maryland, https://www.cs.umd.edu/~sfeizi/}\\
  Department of Computer Science\\
  University of Maryland\\
  College Park, MD 20742\\
  \texttt{sfeizi@cs.umd.edu}\\
}

\begin{document}


\maketitle

\begin{abstract}
Generating high-quality and interpretable adversarial examples in the text domain is a much more daunting task than it is in the image domain. This is due partly to the discrete nature of text, partly to the problem of ensuring that the adversarial examples are still probable and interpretable, and partly to the problem of maintaining label invariance under input perturbations. In order to address some of these challenges, we introduce {\it sparse projected gradient descent} (SPGD), a new approach to crafting interpretable adversarial examples for text. SPGD imposes a directional regularization constraint on input perturbations by projecting them onto the directions to nearby word embeddings with highest cosine similarities. This constraint ensures that perturbations move each word embedding in an interpretable direction (i.e., towards another nearby word embedding). Moreover, SPGD imposes a sparsity constraint on perturbations at the sentence level by ignoring word-embedding perturbations whose norms are below a certain threshold. This constraint ensures that our method changes only a few words per sequence, leading to higher quality adversarial examples. Our experiments with the IMDB movie review dataset show that the proposed SPGD method improves adversarial example interpretability and likelihood (evaluated by average per-word perplexity) compared to state-of-the-art methods, while suffering little to no loss in training performance.
\end{abstract}

\section{Introduction}
Adversarial training (\cite{goodfellow2015fsgm}) is an approach to training neural classification models that involves augmenting the training data with adversarial examples in order to regularize the model. This form of regularization has been shown to be particularly effective in the text domain, providing considerable increases in test accuracy over non-adversarially trained models (\cite{miyato2017adversarial}, \cite{sato2018interp}). Various approaches to adversarial training for text have also been studied for their utility in analyzing and debugging text classifiers (\cite{ribeiro2018semantically}). This application is particularly important, since it has been shown that practitioners consistently overestimate their models' generalization to unseen data (\cite{patel2008investigating}), a problem compounded by text classifiers' tendency toward over-sensitivity and by the enormous variety of text-in-the-wild\,---\,user sentences that diverge only slightly from inputs seen during training can often cause text classifiers to emit incorrect labels (\cite{jia2017adversarial}). This over-sensitivity, conversely, helps to explain the effectiveness of adversarial text training as a regularizer.

Adversarial training requires crafting adversarial examples at train time and training the model jointly on both oiriginal and adversarial examples. In the image domain, these adversarial examples $(\bm{x}', y)$ are often crafted by perturbing ground-truth examples $(\bm{x}, y) \in \mathcal{S}$ through the addition of a small amount of noise $\bm{d}$ with a bounded norm to $\bm{x}$. If an adversarial example is crafted successfully, given $(\bm{x}, y)$ and $\bm{x}' = \bm{x} + d$, the classifier emits a label $y' \neq y$. The model's weights are subsequently adjusted according to the gradient of the joint loss over both $(\bm{x}, y)$ and $(\bm{x}', y)$\,---\,consequently, we would like to ensure both that $\bm{x}'$ is sufficiently similar to other examples in $\mathcal{S}$, and that in perturbing $\bm{x}$ the adversary has not inadvertently changed the ground-truth label from $y$ to some $y' \neq y$ (we call this phenomenon label-inversion in the binary classification setting).

When the input space is not continuous (as the space of all images) but discrete (as the set of all well-formed strings in some language $L$, such as English), it is no longer immediately clear how to add noise to ground-truth examples, much less what a small amount of it might be. A number of solutions to this problem have been proposed, from discrete iterative methods (\cite{papernot2016rnninput}) to methods that perturb a continuous global representation of the sequence (\cite{zhao2018generating}, \cite{iyyer2018paraphrase}). These approaches have their own set of advantages and disadvantages, and we discuss them further in the related work section. 

However, a third class of method, proposed recently in \cite{miyato2017adversarial}, recommends applying perturbations not to sequences of discrete words, but to sequences of their continuous embeddings (\cite{bengio2003neural}). This method of adversarial training for text (here referred to as vanilla AdvT-Text) has been shown to be an excellent regularizer, achieving test accuracy just below contemporary state of the art performance on the IMDB classification task (reported in \cite{johnson2016supervised}). However, it still suffers from a number of issues.
\begin{itemize}
    \item \textbf{Interpretability}. There is no natural way to interpret (or discretize) the perturbed embeddings produced by vanilla AdvT-Text; this makes it difficult to \textbf{(1)} interpret the action of the adversary, \textbf{(2)} to diagnose the reason for the classifier's failure (useful in debugging models), and \textbf{(3)} to assess heuristically whether a label-inversion has taken place.
    \item \textbf{Likelihood}. The perturbed sequences tend to be significantly less probable (as measured by a language model trained on the original dataset) than the ground-truth sequences.
    \item \textbf{Label-Invariance}. Vanilla AdvT-Text perturbs every word embedding in a given sequence, possibly increasing the chance of inverting the ground-truth label.
\end{itemize}
\cite{sato2018interp} introduced a slight modification to vanilla AdvT-Text in an attempt to ameliorate the interpretability issue. This new method, iAdvT-Text, constrains a word embedding's perturbations to lie within the subspace spanned by unit direction vectors to its $K$ nearest neighbors (where $K$ is a hyperparameter), and yields test error similar to vanilla AdvT-Text. Since the resulting adversarial perturbation is a linear combination of unit direction vectors, \cite{sato2018interp} suggests the adversary may be interpreted as moving a word in the direction of the nearest neighbor with highest weight. We find, however, that iAdvT-Text only exacerbates the probability issue, producing sequences with even lower probability than vanilla AdvT-Text; we conjecture this is because of the lack of sparsity constraints on the linear combination's weight vector, as well as the fact that it uses much higher perturbation norms in order to achieve the same performance as AdvT-Text (this is apparent in Figs.~\ref{fig:hmap}, 3, and 4).

\begin{figure}[!t]
    \centering
    \includegraphics[width=1.0\textwidth]{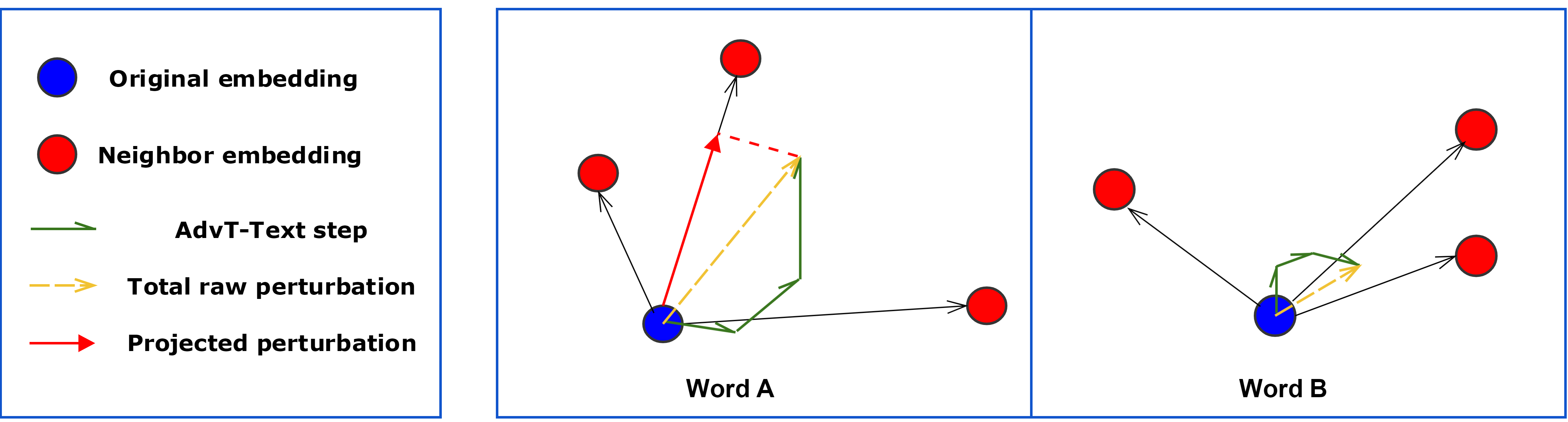}
    
    \justifying
        In this simple two-word example, the adversary takes 3 steps (green arrows) away from the original word, ending at the raw perturbations indicated in yellow. With a sparsity coefficient of $\sigma=0.5$, only the 50\% of perturbations with largest norm are kept\,---\,here, Word A. Finally, Word A's perturbation is projected onto the nearest-neighbor path with the highest cosine similarity, yielding the red arrow.
    \caption{\textit{A simple illustration of SPGD on a two-word sequence.}}
    
    \label{fig:spgd}
\end{figure}

To address these issues, we propose a novel fast-gradient method for adversarial training which we call \textit{sparsified projected gradient descent} (SPGD). Our method essentially adds two steps to the algorithms considered in \cite{miyato2017adversarial} and \cite{sato2018interp} (\textit{v.} Fig.~\ref{fig:spgd} for an illustration):
\begin{itemize}
\item \textbf{Sparsification}. The perturbations computed by vanilla AdvT-Text are sparsified at the sequence level: only the perturbations with highest L2 norms are selected; the rest are zeroed out. This significantly improves the likelihood of the generated adversarial examples, as well as the grammaticality and legibility of their discretizations (\textit{v.} Fig.~\ref{fig:hmap}).

\item \textbf{Directional Projection}. For each embedding, unit-direction vectors to the top-$K$ nearest neighbors are computed. Each high-norm perturbation that survives sparsification is projected onto the nearest-neighbor vector with which it has highest cosine similarity.
\end{itemize}

Thus, unlike \cite{miyato2017adversarial} and \cite{sato2018interp}, our method only perturbs words that contribute significantly to the correct label's probability density, and those words it does perturb, it moves towards at most \textit{one} of its nearest neighbors. In experiments with the IMDB movie review corpus, we find that our training method sacrifices little to no final classifier performance and produces perturbed sequences with significantly higher probability under a language model (LM) trained on the same corpus (see Table~\ref{tab:perp}, below; recall that lower perplexity is better).

\begin{table}[!h]
    \caption{Adversarial  Quality}
    \centering
    \begin{tabular}{@{}p{0.35\textwidth}*{3}{L{\dimexpr0.22\textwidth-2\tabcolsep\relax}}@{}}
        \toprule
        \textbf{Attack} & \textbf{LM Test Perplexity} & \textbf{Avg. Perplexity Gap} \\
        \midrule
        \textit{None (ground-truth corpus)}   & \textit{117.93} & \textit{0.0}\\
        \addlinespace[1mm]
        \midrule
        Vanilla AdvT-Text & 122.53 & 4.60\\
        \midrule
        iAdvT-Text & 123.86 & 5.93\\
        \midrule
        SPGD & \textbf{119.02} & \textbf{1.09}\\
        \bottomrule
    \end{tabular}
    \label{tab:perp}
\end{table}

In summary, our contributions in the present paper are:
\begin{itemize}
    \item We introduce a novel, more interpretable adversarial attack that moves word embeddings toward one (and one only) nearby word
    \item We show a natural adaptation of perturbation sparsity constraints (often found in the image domain) to the text domain
    \item We show that the highly constrained, sparse perturbations generated by SPGD have a similar regularizing effect to that of previous, less interpretable fast gradient methods while improving the linguistic quality of adversarial examples
\end{itemize}

\section{Related Work}
The phenomenon of adversarial examples was first noted in \cite{szegedy2014intriguing}. Thereafter, \cite{goodfellow2015fsgm} proposed injecting adversarial examples into a model's training data in order to regularize the model and improve test generalization (referred to here as AdvT). Early work in adversarial examples focused on the image domain, where example inputs have a natural continuous representation. Examples in the text domain, on the other hand, present peculiar difficulties due to their discrete nature; the sorts of methods devised to contend with these difficulties fall into three broad classes, which we address in turn, occasionally identifying a method's origins in image-domain work.

\paragraph{Fast gradient methods}
A family of fast-gradient sign methods was introduced in \cite{goodfellow2015fsgm}, and applied to the image domain. A Jacobian-based saliency map attack
(JSMA) was introduced by \cite{papernot2016limitations} and \cite{papernot2017blackbox}, whereby a saliency map computed from the classifier's Jacobian is used to rank input components; thereafter, input components are perturbed iteratively, in order of salience, until the classifier's prediction changes. The method was never applied to text. \cite{kurakin2016adversarial} seems to have first proposed repeated application of FGMS to improve the chance of fooling the classifier.

With \cite{miyato2017adversarial}, fast gradient AdvT was first applied to text classification models at the word embedding level. In contrast to \cite{goodfellow2015fsgm}'s family of FGSM attacks, which only use the gradient sign, \cite{miyato2017adversarial} uses the raw gradient.
\cite{sato2018interp} extends \cite{miyato2017adversarial}'s work, modifying their method to improve interpretability
without sacrificing test accuracy or computational efficiency. Our method (SPGD) follows this line of work, further modifying fast gradient text methods by introducing sequence-level sparsification and projecting\footnote{The idea of using some form of PGD to find adversarial examples in the image domain was recently explored in \cite{madry2017towards}, though our application is, to the best of our knowledge, unique.} onto a measure-zero subset of embedding space.

\paragraph{Global gradient methods}
In contrast to the fast-gradient methods just described, global gradient methods use the model gradient to perturb a global embedding of the entire example. For instance, \cite{zhao2018generating} uses an adversarially regularized autoencoder (\cite{zhao2018arae}) to learn a continuous projection of example sequences; in the adversarial regime, they perturb this global representation, subsequently decoding the perturbed point using an LSTM. \cite{iyyer2018paraphrase} employs a similar approach, using syntactically-controlled paraphrase networks (SCPNs) to generate semantically similar, but syntactically divergent adversarial examples from original, ground-truth examples. The point of these sorts of approaches is usually to generate ``natural" adversarial examples which, unlike those produced by the fast-gradient methods above, may diverge in word order or sentence structure from the original example. Thus, such methods align with\,---\,and represent an alternate approach to\,---\,our goal of generating syntactically well-formed and semantically consistent examples in order to improve model regularization. Unfortunately, text autoencoders tend to suffer from high reconstruction error when dealing with very long sequences (such as those in the IMDB corpus, where sequences can range over 2,000 words), and so global gradient methods cannot, in general, be applied to the same datasets.

\paragraph{Non-gradient, discrete methods}
Most of the past approaches to adversarial text generation, however, do not use the model gradient at all, instead working directly on the (discrete) text input. The earliest of these is due to \cite{papernot2016rnninput}, which proposes iteratively substituting words in a sentence with nearby neighbors until the classifier's label prediction changes; \cite{kuleshov2018adversarial}, unpublished, pursues a similar methodology. Still other recent methods (\cite{belinkov2017noise}, \cite{ebrahimi2017hotflip}, and \cite{li2016erasure}, for example) attack text examples by scrambling, misspelling, or erasing words, or even by introducing out-of-vocabulary (OOV) words; these approaches can allow for more easily debugging and regularizing models in the text domain, but they also suffer from the aforementioned issues of harming syntactic coherence or destroying semantic equivalence between original example and adversarial example. Other rule-based methods, such as \cite{ribeiro2018semantically}, attempt to use hand-crafted rules to mitigate these issues and preserve semantic entailment between adversarially-generated and original example.

\paragraph{Further Reading} Even this discussion is necessarily limited in scope and comprehensiveness, focusing only on the papers we feel are most closely related to the present work. For more in-depth surveys of recent methodologies in adversarial training for text, see \cite{warde2016adversarial, wang2019robustsurvey, gong2018gradtextsurvey, wei2019adversarialtextsurvey}.

\section{Adversarial Training for Text}
We start with a set of examples $\{(\bm{x}^{(j)}, y^{(j)})\}_{j=1}^N$, where each $\bm{x}$ is a sequence of words, and each $y$ is a label. We wish to train a classifier $p(y | \bm{x}; \bm{\theta})$ by finding weights $\bm{\theta}^*$ that minimize negative log-likelihood (NLL):
\begin{align}
    \bm{\theta}^* = \argmin_{\bm{\theta}}\mathcal{L}(\bm{\theta}),\:where\:\mathcal{L}(\bm{\theta})\triangleq\frac{1}{N}\sum_{j=1}^N -\log(y^{(j)}|\bm{x}^{(j)};\bm{\theta})
\end{align}
We denote an example sequence of $T$ words as $\bm{x} = \langle w_i \rangle_{i=1}^T$, where each $w_i$ is a word in vocabulary $\mathcal{W}$. Most of the time, however, we work with a word's continuous embedding. To this end, we define a dictionary $\bm{V} \in \mathbb{R}^{(|V| + 1) \times D}$ of embeddings, where $\bm{V}$ is a matrix. Each row $\bm{v}_i$ of $\bm{V}$ represents the embedding of the $i^{th}$ word in the vocabulary, and $\bm{v}_{|V| + 1}$ is the end-of-sequence marker, \texttt{<eos>}.

\paragraph{Vanilla Adversarial Training for Text}
AdvT-Text (\cite{miyato2017adversarial}) merely introduces an extra term into the objective function:
\begin{align}
    \bm{\theta}^* = \argmin_{\bm{\theta}} \left ( \mathcal{L}(\bm{\theta}) + \lambda\mathcal{L}_{Adv}(\bm{\theta}) \right )
\end{align}
where $\lambda$ is a hyperparameter that interpolates between the adversarial and non-adversarial loss, and $\mathcal{L}_{Adv}$ is defined as follows:
\begin{align}
    \mathcal{L}_{Adv} = \frac{1}{N} \sum_{j=1}^N -\log p(y^{(j)} |\; \bm{x}^{(j)} + \bm{d}^{(j)}; \bm{\theta})
\end{align}
Here $\bm{d}^{(j)}$ is a list of worst-case perturbations added to each embedding\,---\,perturbations that increase classifier loss on this example as much as possible\,---\,so it is ideally the solution to the optimization problem:
\begin{align}
    \bm{d}^{(j)*} = \argmax_{\bm{d^{(j)}},\; \|\mathbf{d^{(i)}}\| < \epsilon} -\log p(y^{(j)} |\; \bm{x}^{(j)} + \bm{d}^{(j)}; \bm{\hat{\theta}})
\end{align}
The constraint $\epsilon$ is introduced to prevent the perturbation from moving too far from the original embedding at risk of inverting the ground-truth label. Here $\bm{\hat{\theta}}$ is a constant copy of the current model weights; we have used the hat to indicate that the stop-gradient function should be applied. Solving this problem exactly is intractable for most interesting models like neural networks, so \cite{miyato2017adversarial} (following \cite{goodfellow2015fsgm}) linearizes $-\log p(y^{(j)} |\; \bm{x}^{(j)} + \bm{d}^{(j)}; \bm{\hat{\theta}})$ around $\bm{x}^{(j)}$, yielding the closed-form approximation
\begin{align}
    \bm{d}^{(j)} = - \epsilon \frac{\mathbf{g}}{\|\mathbf{g}\|},\:where\:\bm{g} = \nabla_{\bm{x}^{(j)}} \log p(y^{(j)} |\; \bm{x}^{(i)}; \bm{\hat{\theta}})
\end{align}
Unfortunately, the perturbed embeddings obtained in AdvT-Text typically no longer correspond to words in the embedding dictionary, so it is difficult to interpret the action of the adversary or to explain its regularizing effect.

\begin{figure}[t]
    \centering
    \subfloat[AdvT-Text]{
        \includegraphics[clip,width=\columnwidth]{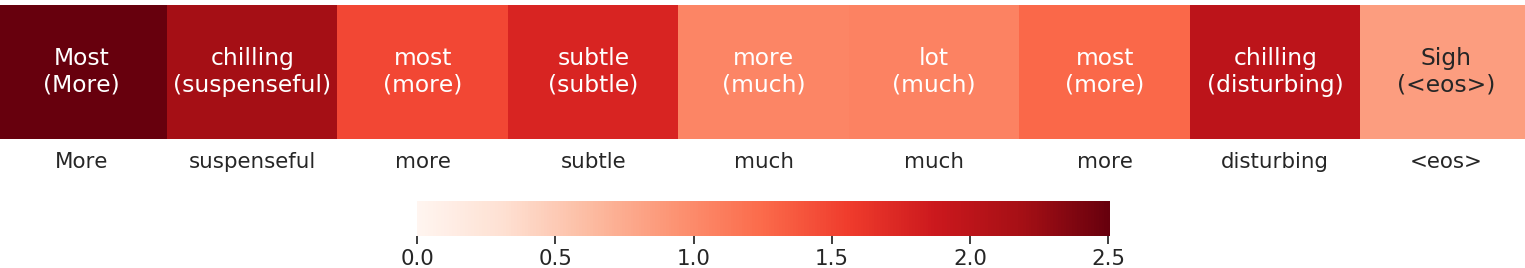}
    } \\
    \subfloat[iAdvT-Text]{
        \includegraphics[clip,width=\columnwidth]{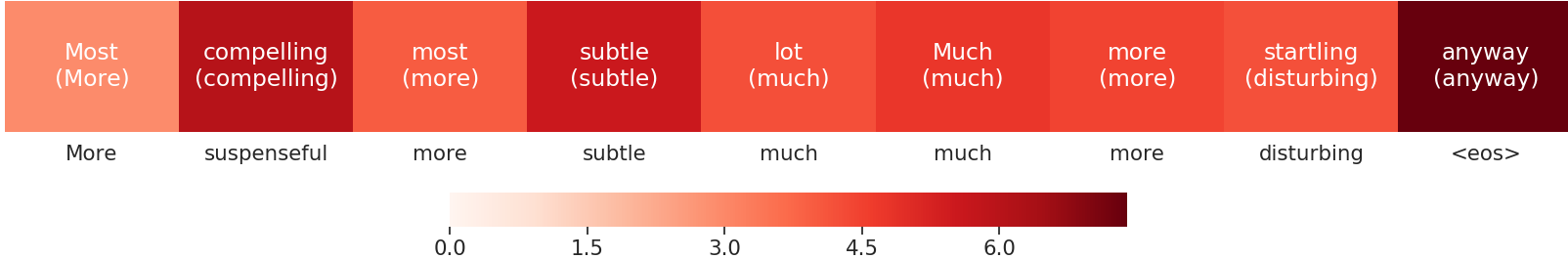}
    } \\
    \subfloat[SPGD]{
        \includegraphics[clip,width=\columnwidth]{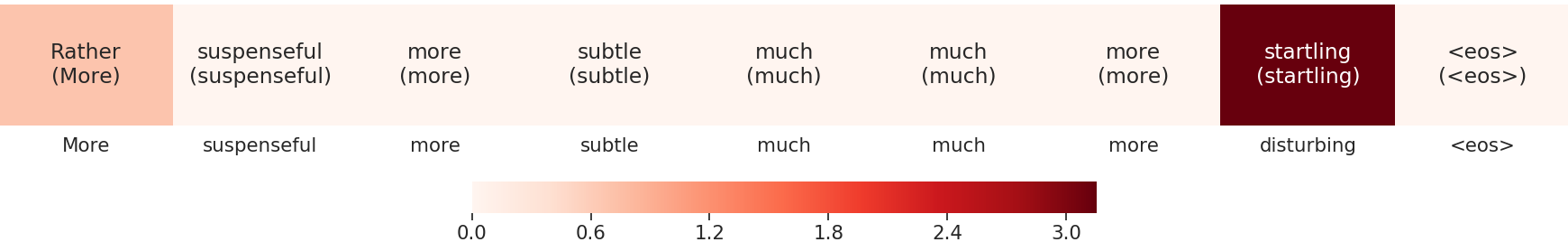}
    }
    \caption{A visual comparison of adversarial sequences, as generated during three training runs using AdvT-Text, iAdvT-Text, and SPGD. SPGD's perturbations are higher quality, preserving the syntax and semantics of the original sequence (for a fuller explanation of the figure, see the Results section).}
    \label{fig:hmap}
\end{figure}

\paragraph{Interpretable Adversarial Training}
This led to the introduction by \cite{sato2018interp} of a more interpretable adversarial training method, iAdvT-Text. In this method, the unit direction vectors are calculated from each original word embedding $\bm{v}_i$ to its $K$ nearest neighbors\,---\,call this set of unit vectors $U_i^K$. Then, iAdvT-Text constrains the perturbation vector for each word embedding in a sequence to lie in the subspace spanned by the vectors in $U_i^K$. In other words, $\bm{d}$ is a linear combination $\sum_{k=1}^K \alpha_k \bm{u}_k^{K}$ for some vector $\bm{\alpha}$ of weights, which is then calculated much as $\bm{d}$ before:
\begin{align}
    \bm{\alpha}^* = - \epsilon \frac{\mathbf{g}}{\|\mathbf{g}\|},\:where\:\mathbf{g} = \nabla_{\bm{\alpha}} \log p_c(y^{(i)} |\; \mathbf{w}^{(i)} + \mathbf{d}; \bm{\hat{\theta}})
\end{align}
However, iAdvT-Text places no constraints on the weight vector, $\bm{\alpha}$, so the adversary can move embeddings towards \textit{or} away from neighbors. Furthermore, the perturbed sequences are easily distinguished from ground-truth sequences by virtue of having significantly higher perplexity than the originals. Note that AdvT-Text suffers from the same issue, though to a slightly lesser degree.

\section{Sparsified Projected Gradient Descent}

\begin{algorithm}[t]
    \caption{The SPGD adversarial training algorithm}
    
    \SetKwInOut{Input}{Input}
    \SetKwInOut{Output}{Output}
    \SetKwProg{SPGD}{SPGD}{}{}
    
    \SPGD{($\bm{w}$)}{
        \Input{\:embedded sequence $\bm{w} = \langle \bm{w}_1, \bm{w}_2, \dots, \bm{w}_N \rangle$\\unit nearest neighbor directions $U_i^K$ for each $\bm{w}_i \in \bm{w}$}
        \Output{\:perturbation vectors $\bm{d} = \langle \bm{d}_1, \bm{d}_2, \dots, \bm{d}_N \rangle$}
        
        \tcp{Calculate raw perturbations $\bm{r}$}
        $\bm{r} \gets \texttt{zeros}(\bm{w}.shape)$ 
        
        \For{$i \in [1, \dots, len(\bm{w})]$}{
            $\bm{r}_i \gets \epsilon \frac{\bm{g}_i}{\|\bm{g}_i\|},\:where\:\bm{g}_i = \nabla_{\bm{w}_i}p_c(y | \bm{w})$ \tcp*{this is just vanilla AdvT-Text}
        }
        
        \tcp{Project perturbations onto nearest neighbor directions}
        $\bm{\hat{r}} \gets \bm{r}$\;
        \For{$i \in [1, \dots, len(\bm{r})]$}{
            $\bm{\hat{r}}_i \gets \left(\bm{r}_i \cdot \bm{u}_i^* \right) \bm{u}_i^*,\:where\:\bm{u}_i^* = \argmax \left( \bm{r}_i \cdot \bm{u}_i^{(j)} \right)$ for $\bm{u}_i^{(j)} \in U_i^K$\tcp*{project}
        }
        
        \tcp{Sparsify perturbations}
        \texttt{K} $\gets \floor{(1 - \sigma)\;N}$,\:\:$\bm{d} \gets \bm{\hat{r}}$\;
        \texttt{TopKIdx} $\gets$ \texttt{TopK}$($\texttt{K}$, \left[\|\bm{r}_i\|_2^2\;for\;each\;i\right])$\;
        \For{$i \in [1, \dots, len(\bm{r})]$}{
            \If{$i \notin$ \texttt{TopKIdx}}{
                $\bm{\hat{r}}_i \gets \bm{0}$\tcp*{zero indices with small perturbations}
            }
        }
    }
    \label{alg:spgd}
\end{algorithm}

In the image domain, adversaries often ensure the quality of the perturbed examples they generate by constraining the perturbation vector's norm; this ensures that adversarial examples remain close to the data-generating distribution. Occasionally, the perturbation vector is sparsified, often by selecting an appropriate $p$-norm (e.g., the $\ell_0$ norm in \cite{shafahi2018adversarial}); \cite{su2019one} even proposes a single-pixel attack. SPGD extends these sparsified attacks into the natural language domain in such a way as to improve adversarial example quality while preserving favorable effects on adversarial training.

SPGD works by applying vanilla AdvT-Text $M$ times to produce a raw candidate perturbation $\bm{r}_i$ for each embedding in a sequence (similarly to \cite{kurakin2016adversarial}, where repeated application of FGSM was found to increase the chance of fooling the classifier). To force a clear interpretation upon each $i^{th}$ raw perturbation, we project it onto the unit neighbor-direction vector in $U_i^K$ with which is has the highest cosine similarity,\footnote{Hence, our method may be considered a form of projected gradient descent (PGD) that projects perturbation gradients onto the set $\mathcal{S}$ consisting of multiples of nearest-neighbor unit directional vectors. $\mathcal{S}$ represents the manipulative power of the adversary, and in the case of SPGD, it is a measure zero subset of embedding space.} yielding
\begin{align}
    \bm{\hat{r}}_i = \left(\bm{r}_i \cdot \bm{u}_i^* \right) \bm{u}_i^*,\:where\:
                \bm{u}_i^* = \argmax_{\bm{u}_i^{(j)} \in U_i^K} \left( \bm{r}_i \cdot \bm{u}_i^{(j)} \right)
\end{align}
Thus, the adversary is constrained to move each word embedding only in a single interpretable direction, towards a neighboring word embedding. Finally, in order to ensure that the adversarial sequence remains similar to the original sequence, we introduce the idea of sparsifying the perturbation vectors at the sequence level. Specifically, given a sparsity factor $\sigma$ (a hyperparameter), we identify the $\floor{(1 - \sigma)\,N}$ perturbation vectors in $\bm{r}$ with smallest L2 norm and set these perturbations to $\bm{0}$. Intuitively, the adversary only uses the perturbations in $\bm{r}$ that are most likely to fool the classifier.

\section{Dataset}
For all of our experiments, we used the well-studied Large Movie Review Dataset benchmark (\cite{maas2011learning}), which consists of $100$K movie reviews from the Internet Movie Database (IMDB). $50$k of the reviews are labeled either positive or negative. To construct a vocabulary $\mathcal{V}$ from the dataset, we remove and ignore words that occur only once (so-called \textit{hapax legomena}), tokenize contractions, remove punctuation, and retain capitalization, for a final vocabulary size of $|\mathcal{V}| = 87,008$. 

\begin{table}[ht]
    \caption{IMDB dataset information}
    \centering
    \begin{tabular}{@{}p{0.25\textwidth}*{5}{L{\dimexpr0.18\textwidth-2\tabcolsep\relax}}@{}}
        \toprule
        & \textbf{Train} & \textbf{Dev} & \textbf{Test} & \textbf{Unlabeled}\tablefootnote{We omit information on the unlabeled sequence lengths because we used them concatenated end-to-end, to train our language model.}\\
        \midrule
        Num. examples & 21,246 & 3,754 & 25,000 & 50,000 \\
        \midrule
        Min. sequence length & 12 & 11 & 7 & \textit{N/A} \\
        \midrule
        Max. sequence length & 2505 & 1101 & 2379 & \textit{N/A} \\
        \midrule
        Avg. sequence length & 242.44 & 239.86 & 235.59 & \textit{N/A} \\
        \bottomrule
    \end{tabular}
    \label{tab:imdb}
\end{table}

See Table~\ref{tab:imdb} (above) for more information on the dataset. Note that the sequences in this dataset are too long to be immediately amenable to the global gradient methods proposed in \cite{zhao2018arae} and \cite{zhao2018generating}.

\section{Models}
For the sake of reproducibility and easy comparison, we adapted the codebase used by Sato et al.~in their \cite{sato2018interp}, making as few changes and additions as possible.\footnote{Sato et al.'s original codebase may be found at \url{https://github.com/aonotas/interpretable-adv}} Like \cite{sato2018interp}, we use a simple, unidirectional LSTM-based classifier (\cite{hochreiter1997long}). We also trained a unidirectional LSTM language model (LM) on examples from the labeled and unlabeled portions of IMDB. The LM was used to initialize the classifier's embedding dictionary and LSTM weights, as well as to rate the quality of adversarial examples (as described in our Results section). All our models use embeddings in $\mathbb{R}^{256}$. While we expect that a bidirectional LSTM (\cite{graves2005framewise}) would have performed significantly better than our uniLSTM, our aim was not to beat current best classifier accuracies\footnote{To the best of our knowledge, the current best test accuracy on this dataset is $95.4\%$, achieved by ULMFiT in 2018 (\cite{howard-ruder-2018-universal}) using a novel approach to inductive transfer learning.}, but merely to compare the effects of various forms of adversarial training. Hence, we opted to use the same style of uniLSTM classifier found in \cite{miyato2017adversarial} and \cite{sato2018interp} to facilitate comparison.

In all, we trained five text classification models on the labeled portion of the IMDB dataset. Two of these were non-adversarially trained baselines, one of which was initialized using our pretrained LSTM language model's weights and embedding dictionary.\footnote{Using a pretrained language model to initialize a text classifier's LSTM yields a significant gain in test accuracy; this effect was first noted and explored in \cite{dai2015semi}, and was employed in both \cite{miyato2017adversarial} and \cite{sato2018interp}.} The other three were likewise initialized using the LSTM language model and trained adversarially using vanilla AdvT-Text (\cite{miyato2017adversarial}), iAdvT-Text (\cite{sato2018interp}), and single-step SPGD (ours). All models, including the LSTM language model, were trained until convergence using the Adam optimizer (\cite{kingma2014adam}) with the standard learning rate.

\section{Results}

Our experiments show that SPGD significantly improves the quality and interpretability of perturbed sequences over vanilla AdvT-Text and iAdvT-Text. At the same time, even given the strict constraints our method imposes on the perturbation calculation, we saw little statistically significant loss in model performance, as measured by test accuracy (reported in Table~\ref{tab:acc}, below).

\begin{table}[ht]
    \caption{Test Results}
    \centering
    \begin{tabular}{@{}p{0.55\textwidth}*{5}{L{\dimexpr0.22\textwidth-2\tabcolsep\relax}}@{}}
        \toprule
        \textbf{Model} & \textbf{Test Accuracy} & \textbf{Test Error Rate}\\
        \midrule
        Baseline  &89.83 &10.17\\
        \midrule
        Pretrained &92.69 &7.31\\
        \midrule
        Pretrained w/AdvT-Text &93.58 &\textbf{6.42}\\
        \midrule
        Pretrained w/iAdvT-Text &93.58 &\textbf{6.42}\\
        \midrule
        Pretrained w/SPGD (1 iter, 75\% sparsity) &93.54 &\textit{6.46} (\textit{ours})\\
        \bottomrule
    \end{tabular}
    \label{tab:acc}
\end{table}

The perturbed sequences produced by SPGD are also amenable to much simpler interpretation: to project them back onto discrete natural language space, one identifies each non-zero perturbation with the nearest neighbor toward which it moved. For example, consider Figure~\ref{fig:hmap}, which compares adversarial examples produced from the same sequence by AdvT-Text, iAdvT-Text and SPGD. The original sequence is represented under each heatmap, and the cell colors represent the magnitude of perturbation. Within each cell, the first word denotes the discretization of the perturbed embedding (here, as in \cite{sato2018interp}, the nearest neighbor towards which the perturbed embedding most nearly moved, as measured by cosine similarity); the second word, in parentheses, denotes the nearest neighbor to each perturbed embedding (the reader will note that optimal hyperparameters for all three methods tend to yield vectors that still lie closer to the original embedding than to any other). The sequences produced by SPGD are very nearly like the original, but with a handful of semantically important words shifted slightly. More importantly, the discretized sequences are relatively grammatical compared to the AdvT-Text and iAdvT-Text examples, and cases where the adversary has clearly inverted the example's label are rare. By contrast, the sequence in Fig.~4 (Appendix A) has been perturbed by AdvT-Text and iAdvText to such a degree that it's difficult to assess in good faith what their new ground-truth label might be; the discretizations of SPGD's adversarial examples, however, remain mostly well-formed and semantically similar to the originals.

We can also compare the distribution of adversarial examples with the data generating distribution. By contrast with AdvT-Text and iAdvT-Text, we find that SPGD's adversarial examples are on average much less distinguishable from the original dataset. We measured this effect using an LSTM language model trained on the entire IMDB corpus, and in Table~\ref{tab:perp} we report our results in terms of average per-word perplexity over a large random sample of the test set. Recall the definition of perplexity:
\begin{align}
    \exp{\left(-\frac{1}{N}\sum_{j=1}^N \log (\bm{x}^{(j)} | y^{(j)})\right)}
\end{align}
As Table~\ref{tab:perp} shows, SPGD narrows the perplexity gap between adversarial and original sequences over other methods. Interestingly, we observed that higher sparsity coefficients ($\sigma \approx 0.75$) yielded better test accuracies. Altogether, we believe these results strongly suggest that in the text domain {\it more realistic adversarial examples regularize better}, a suggestion that we hope will be take into account by future research in the area.

\section{Discussion}
We have presented a novel adversarial perturbation method (SPGD) and demonstrated its utility in adversarial training. Our experiments have shown that SPGD produces higher-quality, more interpretable perturbed sequences than previous fast-gradient methods for text without sacrificing final classifier accuracy. However, while our method addresses the problem of preserving label invariance under perturbation, it addresses it only indirectly by restricting the percentage of embeddings in a sentence that an adversary is allowed to perturb. We suggest future work explore a more direct approach, whereby a class-conditional LSTM $p_l(\bm{x}|y)$ is trained on the dataset and added to the adversarial gradient term. Thus, the computation of $\bm{d}$ in vanilla AdvT-Text becomes:
\begin{align}
    \bm{d} = - \epsilon \frac{\bm{g}}{\|\bm{g}\|},\:where\:\bm{g} = \nabla_{\bm{x}} \left[ \log p(y |\; \bm{x}) - \log p_l(\bm{x} |\; y) \right]
\end{align}

The set of adversarial sequences generated by SPGD and its predecessors represents only a small subset of the set of all possible adversarial sequences: it excludes, for instance, paraphrases and other sequences where the word order or sentence structure has changed, but the meaning (or the label) has remained invariant. Recent work (\cite{zhao2018generating}, \cite{iyyer2018paraphrase}) has attempted to address these restrictions using autoencoders and SCPNs, but such approaches are limited by the ability of latent-variable generative text models to encode and decode very long sequences (such as those in IMDB) with high reconstruction accuracy. More work needs to be done here.

\small

%
%
%



\section*{Appendix A}
Here, and on the following pages, we provide some further randomly selected examples comparing the perturbed sequences generated by the fast-gradient text methods considered in this paper: AdvT-Text, iAdvT-Text, and SPGD.

\vfill
\begin{figure}[h]
    \centering
    \subfloat[AdvT-Text]{
        \includegraphics[clip,width=\columnwidth]{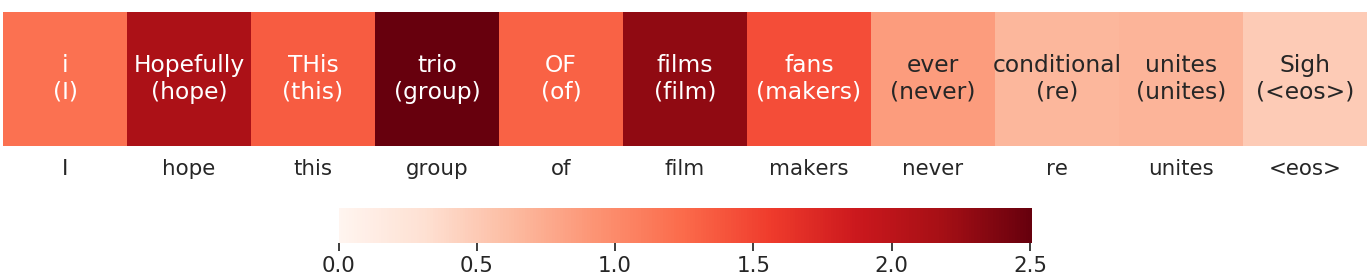}
    } \\
    \subfloat[iAdvT-Text]{
        \includegraphics[clip,width=\columnwidth]{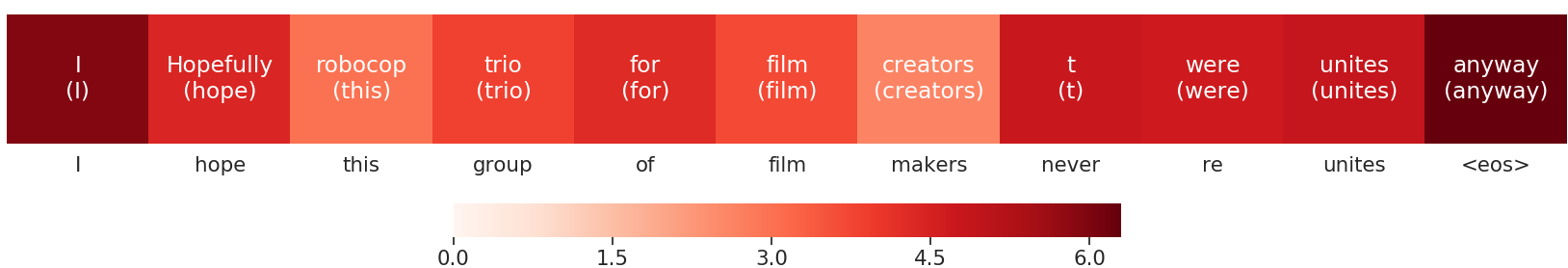}
    } \\
    \subfloat[SPGD]{
        \includegraphics[clip,width=\columnwidth]{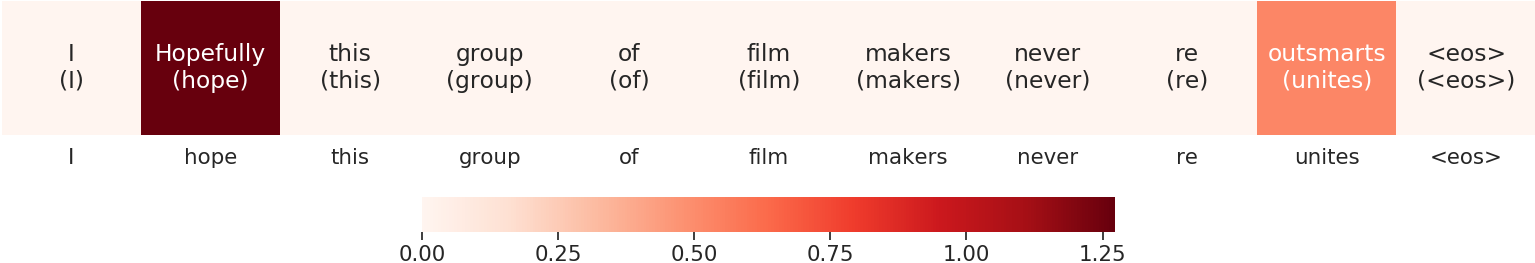}
    }
    \caption{Adversarial sequences generated (from top to bottom) by AdvT-Text, iAdvT-Text, SPGD.}
    \label{fig:hmap-appendix1}
\end{figure}
\vfill

\newpage
\null
\vfill

\begin{figure}[h]
    \centering
    \subfloat[AdvT-Text]{
        \includegraphics[clip,width=\columnwidth]{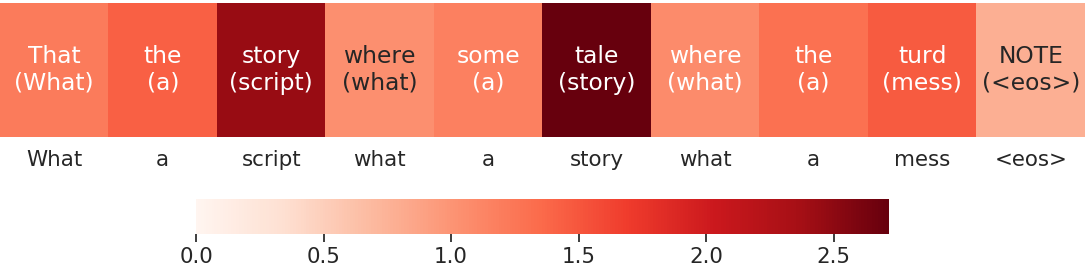}
    } \\
    \subfloat[iAdvT-Text]{
        \includegraphics[clip,width=\columnwidth]{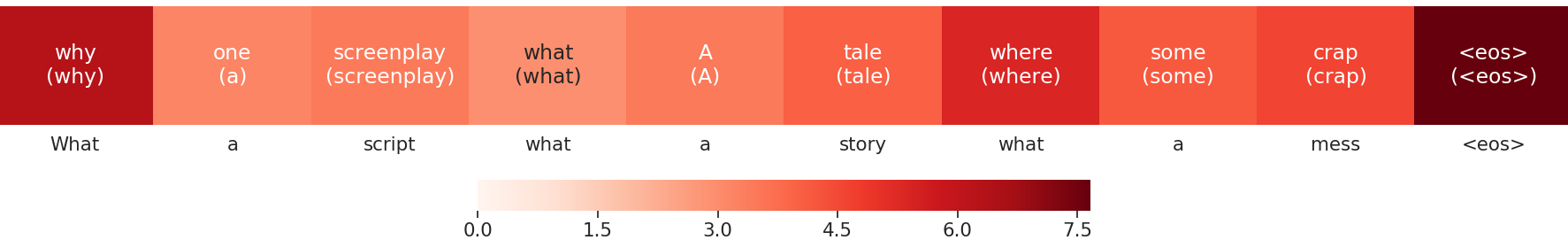}
    } \\
    \subfloat[SPGD]{
        \includegraphics[clip,width=\columnwidth]{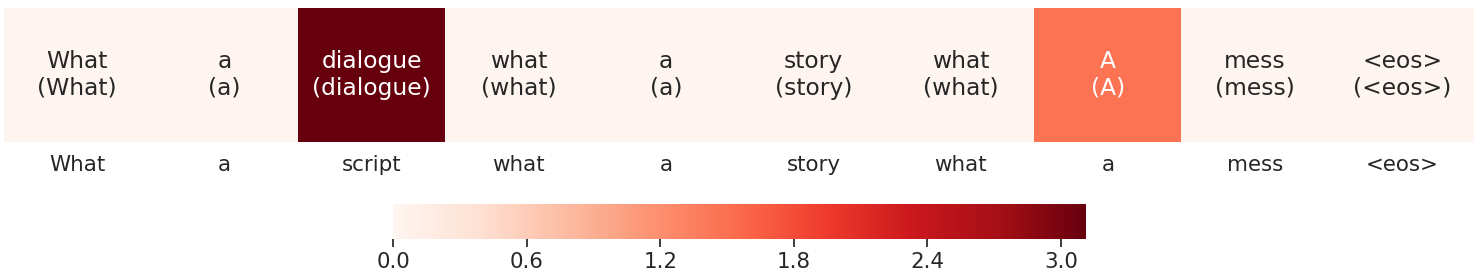}
    }
    \caption{Another set of sequences generated (from top to bottom) by AdvT-Text, iAdvT-Text, SPGD.}
    \label{fig:hmap-appendix2}
\end{figure}
\vfill

\newpage
\section*{Appendix B: Hyperparameters}

Our codebase extends that used by Sato et al.~in their \cite{sato2018interp}, making as few changes and additions as possible. All models were specified and trained using Chainer \cite{tokui2015chainer} (an expedient adopted purely for compatibility with \cite{sato2018interp}'s codebase), and code was written using Python $3.6$.

In all our models, words are first embedded in $\mathbb{R}^{256}$ using an embedding dictionary $\mathcal{E}$ learned jointly with the model weights $\bm{\theta}$. Unless otherwise stated, embeddings were initialized using a diagonal zero-mean, unit-variance Gaussian.

For our classifier (like \cite{sato2018interp}), we use a unidirectional LSTM (\cite{hochreiter1997long}) with hidden dimension $1024$ connected to a hidden feedfoward layer comprising $30$ units with ReLU activation (\cite{nair2010rectified}). The classifier's output layer consists of $2$ feedforward units, to which we apply an adaptive softmax activation (\cite{grave2017efficient}). In the baseline model, weights were initialized independently at random using a scaled Gaussian distribution with parameters
\begin{equation}
    \mu=0,\:and\:\sigma=\sqrt{\frac{1}{fan_{in}}}
\end{equation}
where $fan_{in}$ is the number of units in the layer, as described in \cite{lecun2012efficient} (this is sometimes called \textit{Lecun initialization}).

As mentioned in the paper main body, we use in some of our experiments a unidirectional LSTM language model. The language model (LM), like the classifier, uses embeddings in $\mathbb{R}^{256}$ and an LSTM-unit hidden size of $1024$. The LM weight matrices were initialized uniformly at random to values in $[-0.1, 0.1]$, and its LSTM's forget-gates were initialized to $1.0$. It was trained until convergence using the Adam optimizer (\cite{kingma2014adam}), with the standard learning rate of $\alpha=10^{-3}$, and an $\alpha$-decay rate of $0.9999$, applied whenever validation perplexity failed to fall at the end-of-epoch evaluation, reaching a final test perplexity of $114.77$. As in \cite{sato2018interp}, the dataset used to train the LM was a conjunction of the standard IMDB labeled training data (consisting of $25,000$ movie reviews) and the unlabeled IMDB examples (consisting of $50,000$ movie reviews), all concatenated end to end.

In total, we trained five uniLSTM movie-review classifiers; these are described in the paper main body, and their performance reported in Tables 1 and 3. All model evaluations were run at least $5$ times, with the best results kept for each model. The two baseline models, as well as the AdvT-Text and iAdvT-Text models were trained using the same hyperparameters employed by \cite{sato2018interp}\,---\,in particular, the AdvT-Text model used adversarial step size of $\epsilon=5.0$, and the iAdvT-Text model used a step size of $\epsilon=15.0$ and produced perturbations that were linear combinations of $K=15$ nearest neighbor directional vectors. The last model, trained using SPGD, used an adversarial step of size $\epsilon = 25.0$ projected onto a direction chosen from paths to the top $K=15$ nearest neighbors, using sparsity $\sigma=0.75$ (we found that a higher sparsity coefficient provided better results). We recomputed nearest neighbors every $50$ batches, in order to account for the changing embedding during training. To determine best $K$, $\epsilon$, and $\sigma$, we used an elided grid search, trying combinations of parameters under the assumption that model generalization error was roughly convex in each hyperparameter independently\,---\,this was necessary to conserve time.

To maintain gradient stability in the classifier's LSTM, gradients were clipped to a norm of $4.0$ (as described in \cite{pascanu2013difficulty}). In training the LSTM language model, we found we could clip gradients at $5.0$ and maintain stability.

All training was performed on an NVIDIA Tesla V100 in the 16GB configuration. In order to cope with memory constraints, our LM training used truncated backpropagation through time (TBPTT), first described in \cite[pg.~23]{sutskever2013training}, with a backpropagation length limit of 35.

\end{document}